\documentclass{article}

\usepackage[final]{neurips_2021}

\usepackage[utf8]{inputenc} % allow utf-8 input
\usepackage[T1]{fontenc}    % use 8-bit T1 fonts
\usepackage{hyperref}       % hyperlinks
\usepackage{url}            % simple URL typesetting
\usepackage{booktabs}       % professional-quality tables
\usepackage{amsfonts}       % blackboard math symbols
\usepackage{nicefrac}       % compact symbols for 1/2, etc.
\usepackage{microtype}      % microtypography
\usepackage{xcolor}         % colors
\usepackage{caption}
\usepackage{subcaption}
\usepackage{graphicx}
\usepackage{booktabs}
\usepackage{multirow}
\title{Counterfactual Fairness in Mortgage Lending via Matching and Randomization}

\author{%
  Sama Ghoba\\
  Albers School of Business and Economics\\
  Seattle University\\
  \texttt{sghoba@seattleu.edu} \\
  \And
  Nathan Colaner\\
  Albers School of Business and Economics\\
  Seattle University\\
  \texttt{colanern@seattleu.edu} \\
}

\begin{document}

\maketitle

\begin{abstract}
  Unfairness in mortgage lending has created generational inequality among racial and ethnic groups in the US. Many studies address this problem, but most existing work focuses on correlation-based techniques. In our work, we use the framework of counterfactual fairness to train fair machine learning models. We propose a new causal graph for the variables available in the Home Mortgage Disclosure Act (HMDA) data. We use a matching-based approach instead of the latent variable modeling approach, because the former approach does not rely on any modeling assumptions. Furthermore, matching provides us with counterfactual pairs in which the race variable is isolated. We first demonstrate the unfairness in mortgage approval and interest rates between African-American and non-Hispanic White sub-populations. Then, we show that having balanced data using matching does not guarantee perfect counterfactual fairness of the machine learning models.
\end{abstract}

\section{Introduction}
As dependence on machine learning algorithms increases, fairness concerns about them have also increased \citep{barocas-hardt-narayanan}. To quantify fairness and address algorithmic bias, researchers have proposed many fairness metrics \citep{Kleinbergpandp.20181018,verma2018fairness,mehrabi2021survey}, including both group and individual fairness metrics. Among the individual fairness metrics, the counterfactual fairness metric is one of the most popular because of its simplicity and potential to allows for causal interpretations \citep{hardt2016equality,chouldechova2017fair, kusner2017counterfactual, makhlouf2020survey}. 

Counterfactual fairness not only provides a fairness metric, but also proposes a way to achieve it through causal modeling. For example, in a proposal by \citep{kusner2017counterfactual}, the objective is to learn a Bayesian latent variable representation of key factors in decision making such that the latent variables are independent of the source of the bias (e.g., race or sex). However, this approach has been criticised because of its reliance on additional biases created by the modeling assumptions \citep{lee2021algorithmic}.

To avoid any potential bias created by modeling assumptions \citep{lee2021algorithmic}, we use the causal matching approach \citep{imbens2015causal,rosenbaum1983central,stuart2010matching}, and evaluate the success of the counterfactual fairness framework in learning fair algorithms using the mortgage lending data. Our causal matching creates a balanced dataset in which for each African-American mortgage application, there is a non-Hispanic White application with similar financial standing. Causal matching is transparent and easy to understand for a non-technical audience. Moreover, it provides the counterfactual pairs that make the evaluation of counterfactual fairness less biased. Having matched pairs also allow us to randomly swap the labels within each pair, which guarantees that all biases in the data due to racial factors are removed.

We analyze the 2019 mortgage application data (that was pre-COVID-19) made available through the Home Mortgage Disclosure Act (HMDA). We focus on individual home purchase applications and compare the application approval and interest rates for African-American and non-Hispanic White sub-populations. Because there are fewer African American applications, for every African-American application, we match a non-Hispanic White application with similar financial standing. 

We perform an ablation study with four variations of the algorithm.
The results show that excluding the race variable from the feature set has the largest impact on making the algorithm fairer. Randomization also helps, especially in the interest rate prediction task. Looking at the accuracy of the algorithms in the two racial sub-groups, we show that on the matched data, in which the financial factors are independent of the race, the results are still not perfectly fair. Our empirical finding confirms the hypothesis that fairness is not just a data problem \citep{HOOKER2021100241}. Even in the simplest and yet one of the most consequential data-sets, balancing data using a non-parametric and transparent method such as causal matching does not \textit{perfectly} remove the biases in the algorithms. 

% \noindent\rule{\textwidth}{0.4pt}

% Lee and Floridi (2021) critique several models of algorithmic fairness on the grounds that they impose "absolute mathematical condition[s]" on fairness, and that they "do not account for bias embedded in the data" (178). Both of these problems ultimately undermine algorithmic fairness, on their view. The first problem creates unrealistic limits on actual decision-making, and the second problem creates[??]... 

% They believe that four particular approaches all have these two flaws in common. They identify those approaches as group fairness, equalisation of error metrics, individual fairness, and counterfactual fairness. The critique of counterfactual fairness as an absolute mathematical condition and as a model that does not account for existing bias is not without merit, although it is ultimately off base. Their critique is based on a specific approach to counterfactual fairness. This is the latent variable approach, articulated in Kusher et al (2018) and Loftus et al (2018).

% When their critique is confined to the latent variable approach, it has merit. However, their critique does not fully consider the causal matching approach, which is another valid approach to counterfactual fairness. Using the example of mortgage lending data, we show how causal matching generates a model of algorithmic fairness via counterfactual fairness, and then why the causal matching approach is not problematic in the ways imagined by Lee and Floridi.
\section{HMDA Data Description}
Evaluation of fairness in mortgage lending requires information such as mortgage payment and credit score information that is not available in public data. However, similar to \citep{lee2021algorithmic}, we can still perform comprehensive studies about fairness in mortgage approval and interest rates using the data provided by the Home Mortgage Disclosure Act (HMDA) \footnote{\url{https://ffiec.cfpb.gov/data-browser/data/2020}}.

From the HMDA data, without loss of generality, we focus on two predominant racial groups: African-Americans and non-Hispanic Whites. We also focus on the 2019 data, the last year before the COVID-19 pandemic, to avoid the effects of the pandemic. We further narrow down the data to residential home purchases.

Our data has 3,186,974 unique mortgage applications. Among these are 2,824,322 non-Hispanic White and 362,652 African-American applications. Based on the number of mortgage applications in HMDA data, the B/W ratio is 12.8\%. Note that the ratio of African-Americans to non-Hispanic Whites in the US population is 22.3\%\footnote{\url{https://www.census.gov/quickfacts/fact/table/US/PST045219}}. In our data, the average approval rate is 88.3\%. While this rate is 89.8\% for non-Hispanic White applicants, it is 76.7\% for African-American applicants.

We use the following variables in our analysis: 'interest rate','applicant sex','income','applicant race', 'state code', 'loan type', 'debt to income ratio', 'loan to value ratio', and 'lien status'. We provide a detailed description of these variables in the appendix.

% Follow \citet{lee2021algorithmic} for data description section.

\begin{figure}[t]
     \centering
     \begin{subfigure}[b]{0.49\textwidth}
         \centering
         \includegraphics[width=\textwidth]{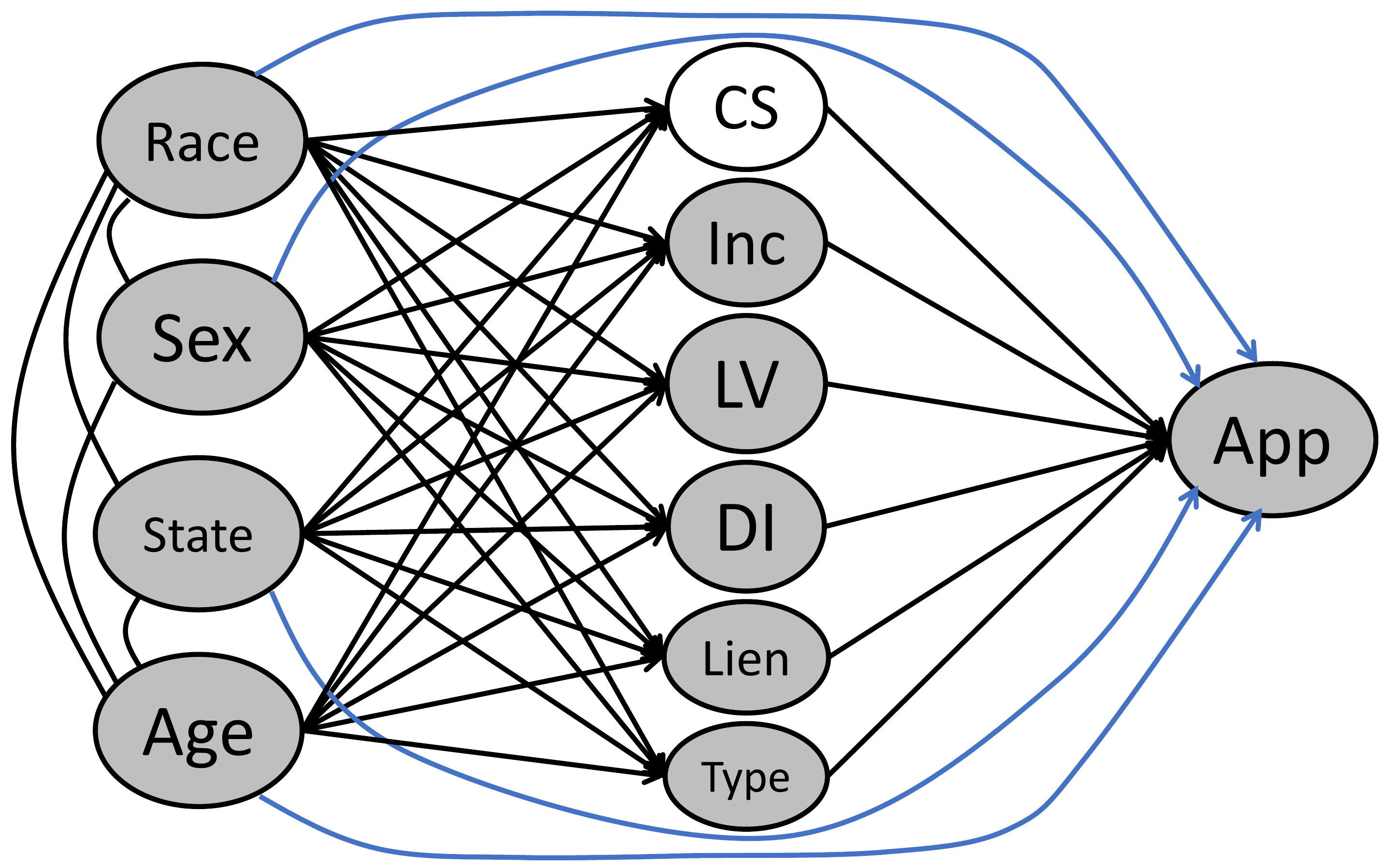}
         \caption{The original graph}
         \label{fig:causal_graph}
     \end{subfigure}
     \hfill
     \begin{subfigure}[b]{0.49\textwidth}
         \centering
         \includegraphics[width=\textwidth]{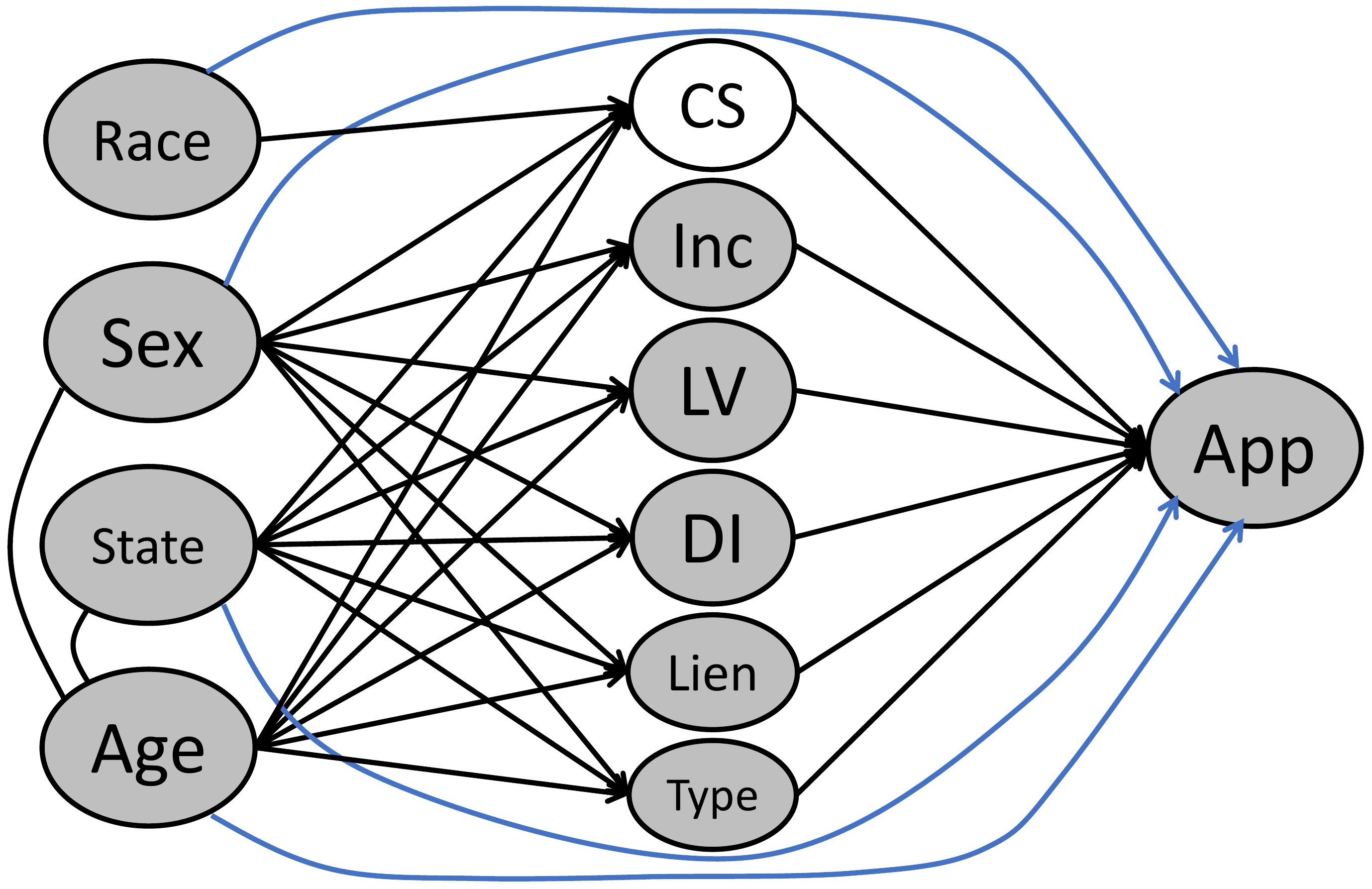}
         \caption{The graph for the matched data}
         \label{fig:graph_matched}
     \end{subfigure}
        \caption{The impact of matching on the graph. It breaks the dependence links between race and the matched variables, except the outcome and unobserved variables. Randomly swapping the labels within each pair breaks the Race $\rightarrow$ CS link. We have used the following abbreviations: CS: Credit Score, Inc: Income, LV: Loan to Value ratio, DI: Debt to Income ratio, App: Loan approval outcome. To avoid clutter, we do not show the latent variables corresponding to each node.}
        \label{fig:causal_graphs}
\end{figure}

\section{Causal Graphs, Matching, and Randomization}
\begin{figure}[h]
     \centering
     \begin{subfigure}[b]{0.49\textwidth}
         \centering
         \includegraphics[width=\textwidth]{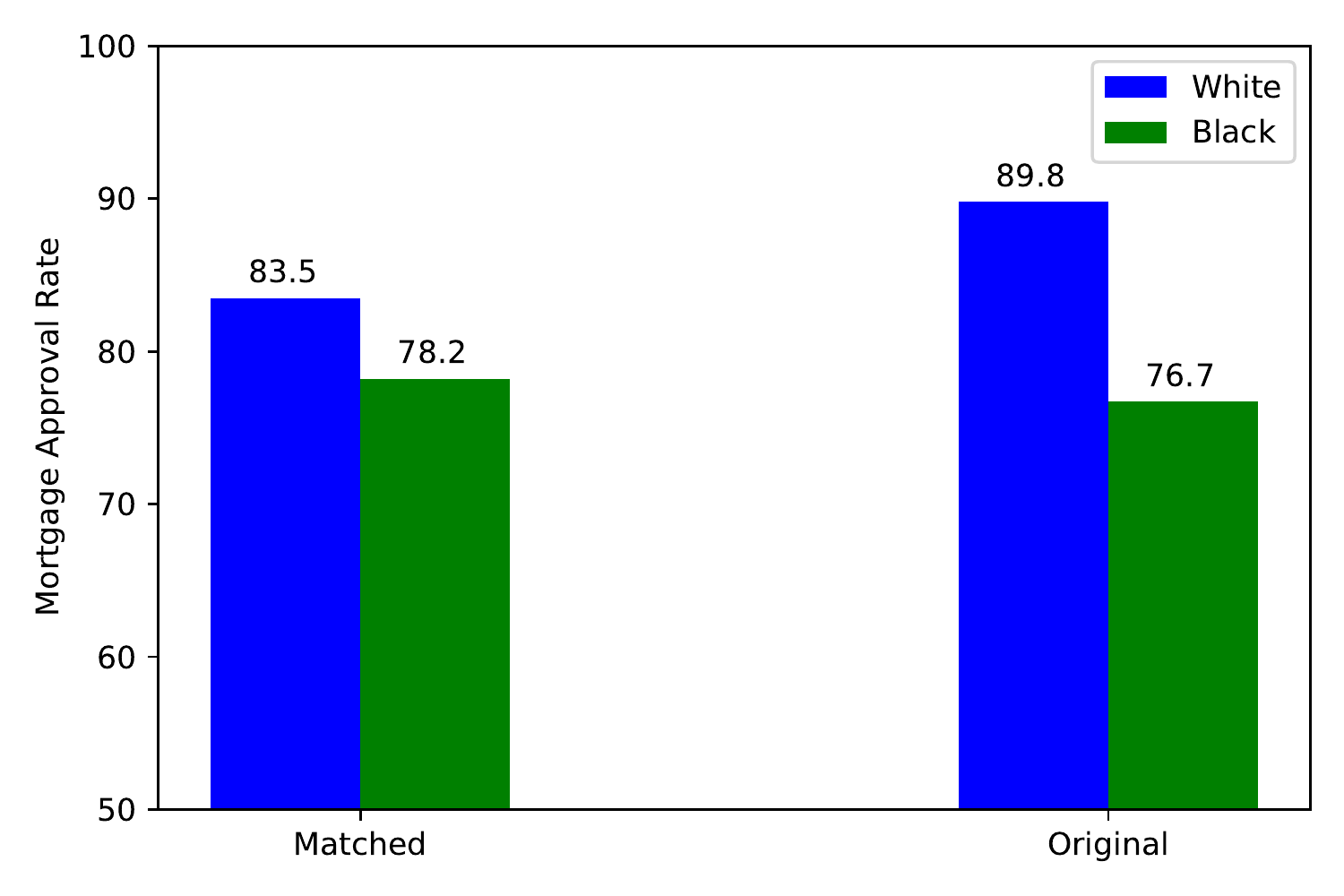}
         \caption{Approval Rate}
         \label{fig:app_prob}
     \end{subfigure}
     \hfill
     \begin{subfigure}[b]{0.49\textwidth}
         \centering
         \includegraphics[width=\textwidth]{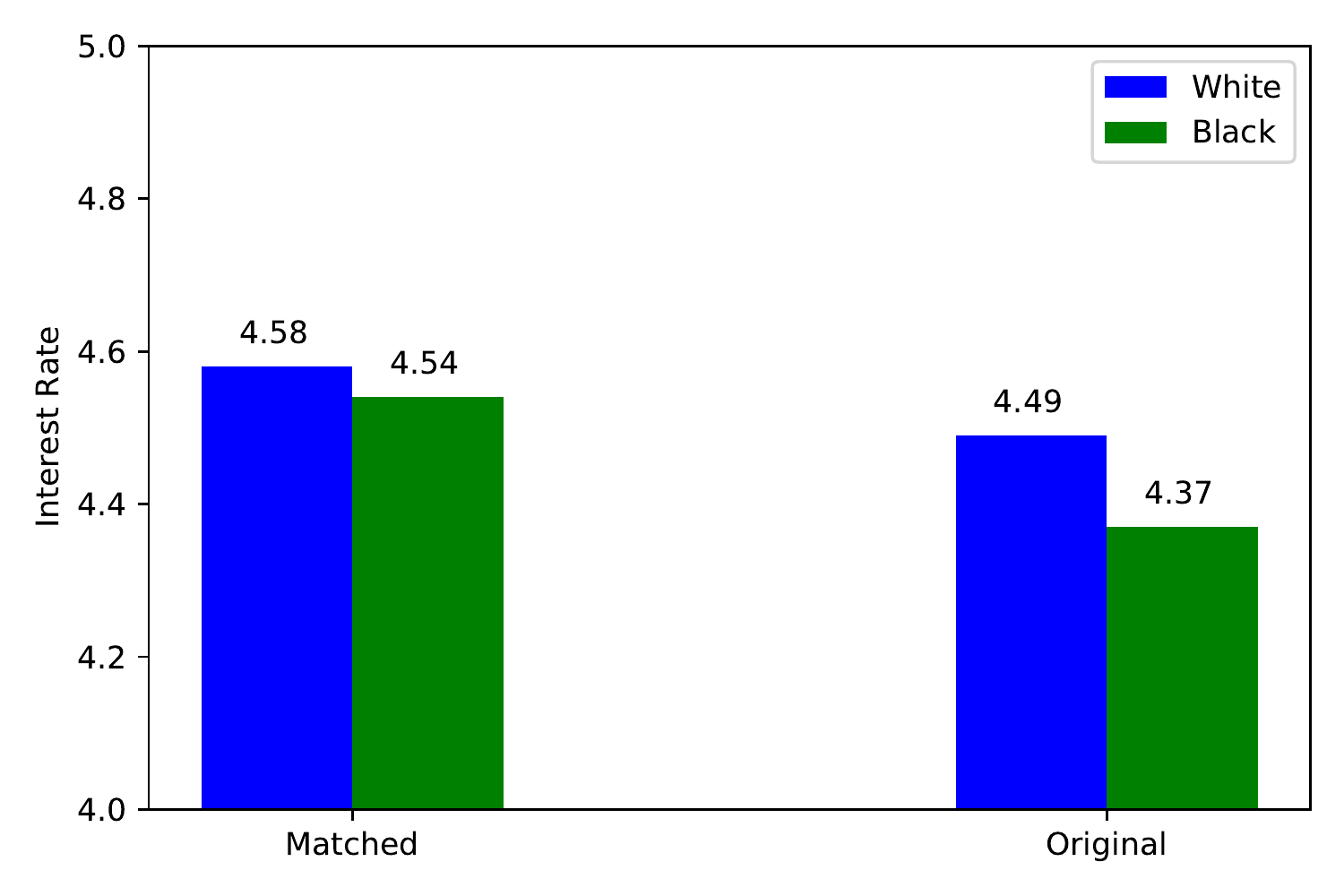}
         \caption{Average Interest Rate}
         \label{fig:int_rate}
     \end{subfigure}
        \caption{The Impact of Matching on Population Statistics.}
        \label{fig:matching_effect}
\end{figure}

We propose the graph in Figure \ref{fig:causal_graph} that shows the relationships among the variables in the data. Note that our graph is not a fully directed acyclic graph (DAG) because we model the correlation between the demographics with undirected edges. The graph shows that demographic variables such as race and sex have impact on the financial status of the applicants. Also, they have an additional direct impact on the change of approval, which means race has both direct and indirect impact.

Figure \ref{fig:graph_matched} shows that by matching we can remove the dependency between the race and other variables. We perform exact matching by binning the real valued variables such as income, loan to value ratio, and age (details in the appendix) \citep{blackwell2009cem}. For each African-American application we find a non-Hispanic White application with the same characteristics in all variables (except race and outcome variables). In the algorithmic fairness literature, our exact matching is different from the nearest neighbor matching used by \citet{khademi2019fairness} and \citet{khademi2020algorithmic}.

We chose exact matches because of the simplicity and transparency of that procedure. While we could use propensity score weighting methods \citep{imbens2015causal,rosenbaum1983central,stuart2010matching}, we prefer the matching procedure because it provides matched pairs that are counterfactuals and make the evaluation of counterfactual fairness easier. Note that we could not match every single African-American application in the data: out of 362,652 applications, 339,849 are matched with applicants from the non-Hispanic White population. During matching, we match applications with similar missing data patterns. After matching, we impute the missing values with the mean of each variable. 

We further verify the correctness of our matching by attempting to predict the race using the seven other features in the matched data. Holding out 25\% of the pairs for testing, a Random Forest trained on the rest of the data has a nearly random AUC result at 55\%. Thus, in our matched data, we do not have a significant proxy for race as a function of other features.

Figure \ref{fig:matching_effect} shows the impact of matching on the approval and interest rate. The figures show that while fixing all other factors reduces the disparity between two racial sub-populations, there are still gaps due to only race. 

Finally, we randomize the training set by randomly swapping the labels (approval or interest rate) in each matched training pair with probability 0.5. Similar to the randomized controlled trials, label randomization breaks the unobserved links between race and the features/label variables. It ensures that in the training set, the features and the label have no association with the race variable. It addresses the concern that we cannot completely remove the impact of the race from the training data. Note that we do not randomize the test set on which we evaluate the algorithms. For other techniques to remove the hidden variable bias in counterfactual fairness see \citep{kilbertus2020sensitivity}.

\section{Investigation of Fairness}
After a brief review of the counterfactual fairness definition, we describe the baselines we use and how we train them. Finally, we show our results and analyze them.

\subsection{Background on Counterfactual Fairness}
The counterfactual definition of fairness considers a model fair if upon intervention and change of the race, its output does not change \citep{kusner2017counterfactual}. The intervention is usually performed using a causal graph \citep{pearl2009causality}, similar to our graphs in Figure \ref{fig:causal_graphs}. The intervened data points become the counterfactuals of the original data points. Given this definition, we notice that the pairs in the matched data are counterfactuals. Thus we can easily compute the \textit{CounterFactual Unfairness} using the matched pairs as follows:
\begin{equation}
    CFU = \frac{1}{n}\sum_{i=1}^n |\widehat{y}_i(b) - \widehat{y}_i(w) |,
    \label{eq:cfair}
\end{equation}
\noindent where in the mortgage approval task, $\widehat{y}_i$ is the probability of approval and in the interest rate prediction, it denotes the predicted interest rate.

\subsection{Baselines}
We perform an ablation study to evaluate the impact of matching and randomization using four algorithms:

\textbf{Algorithm 1 (Common Practice in ML)}: Use all features, including the race, to achieve the highest possible accuracy.

\textbf{Algorithm 2 (No Race)}: Do not use the race variable in the predictions. This is not perfect because the race information can be embedded in the other variables.

\textbf{Algorithm 3 (Matching+No Race)}: Use matching to make other variables independent of the race variable. Then, use all variables except race to predict the target variable (approval or interest rate). 

\textbf{Algorithm 4 (Matching+No Race+Label Randomization)}: Use matching to make other variables independent of the race variable. We also randomly with probability 0.5 swap the labels in each matched training pair. Then, use all variables except race to predict the target variable (approval or interest rate).

\subsection{Training Details}
We randomly hold out 25\% of the matched pairs as test set. We remove the test data points from the original data (pre-matched) to avoid data leak in Algorithms 1 and 2. We tune the hyperparameters of the algorithms using cross-validation. We use the \texttt{scikit-learn} \citep{scikit-learn} implementation for all of the machine learning algorithms. We publicly release our code at  \url{https://github.com/samasky/algorithmic_fairness_matching}.

\begin{table}[t]
    \centering
    \caption{Interest Rate Prediction Results}
    \label{tab:regression}
    \begin{tabular}{@{}l@{}|l|c|c|c|c@{}}
        \toprule
         & Algorithm & Test CFU & Test RMSE & Test RMSE /Whites & Test RMSE /Blacks   \\
        \midrule
         \multirow{3}{*}{Linear Regression}& 
         Algorithm 1&0.1124 &0.910 &0.809 & 1.010\\
         & Algorithm 2&0.0244 &0.911 &0.810 & 1.013\\
         & Algorithm 3&0.0229 &0.865 &0.774 &0.956 \\
         & Algorithm 4 & \textbf{0.0218} & 0.865 &0.774 &0.956\\
        \midrule
        \multirow{3}{*}{Random Forest}&
        Algorithm 1& 0.0317 & 0.772 & 0.696& 0.848 \\
         & Algorithm 2& 0.0316 & 0.772 &0.696 & 0.848\\
         & Algorithm 3& 0.0297 &0.763 & 0.686& 0.839\\
         & Algorithm 4& \textbf{0.0217} & 0.758 & 0.681 & 0.835\\
        \bottomrule
    \end{tabular}
\end{table}

\subsection{Results}
\begin{figure}[t]
     \centering
     \begin{subfigure}[b]{0.45\textwidth}
         \centering
         \includegraphics[width=\textwidth]{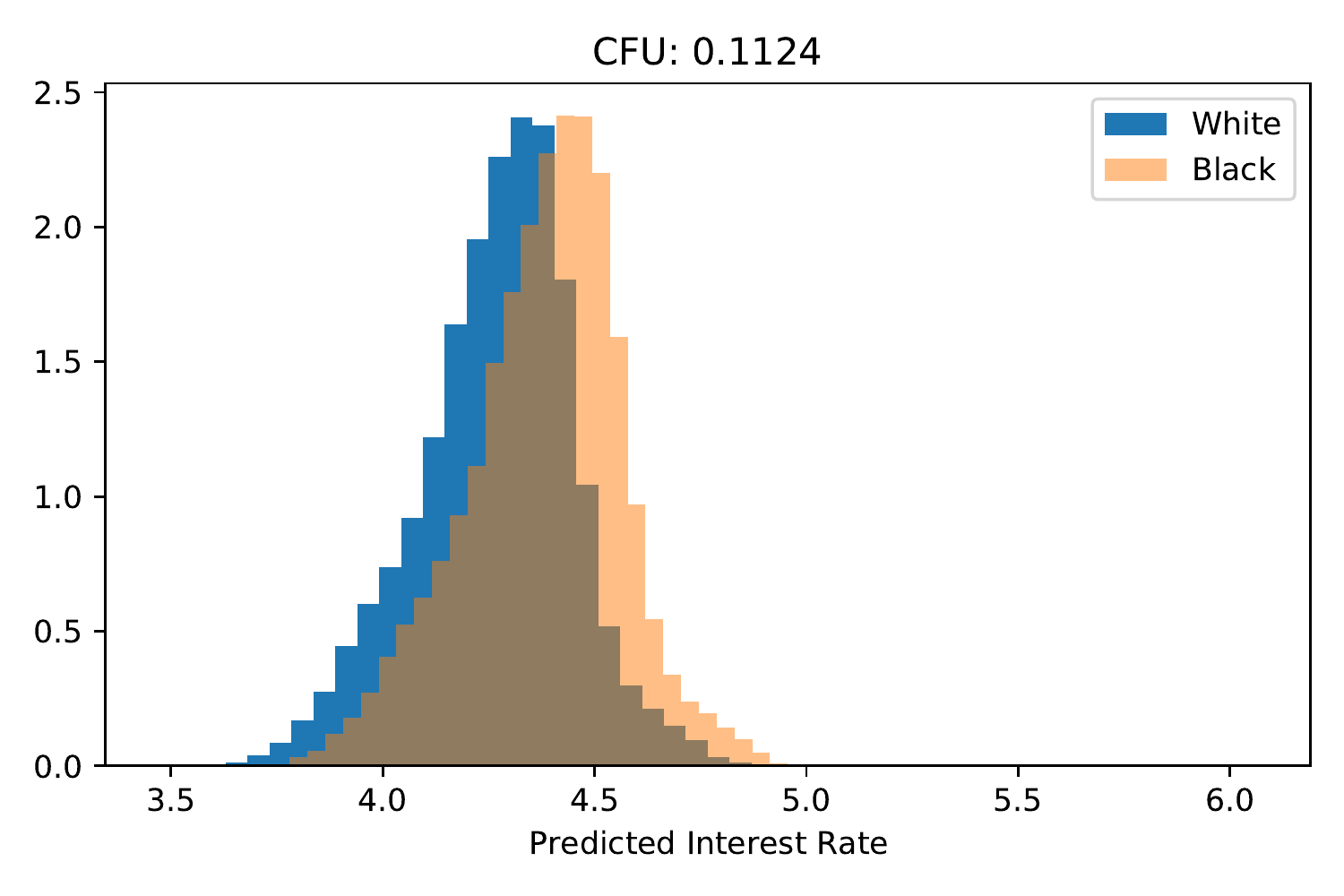}
         \caption{Algorithm 1}
         \label{fig:algo1}
     \end{subfigure}
     \hfill
     \begin{subfigure}[b]{0.45\textwidth}
         \centering
         \includegraphics[width=\textwidth]{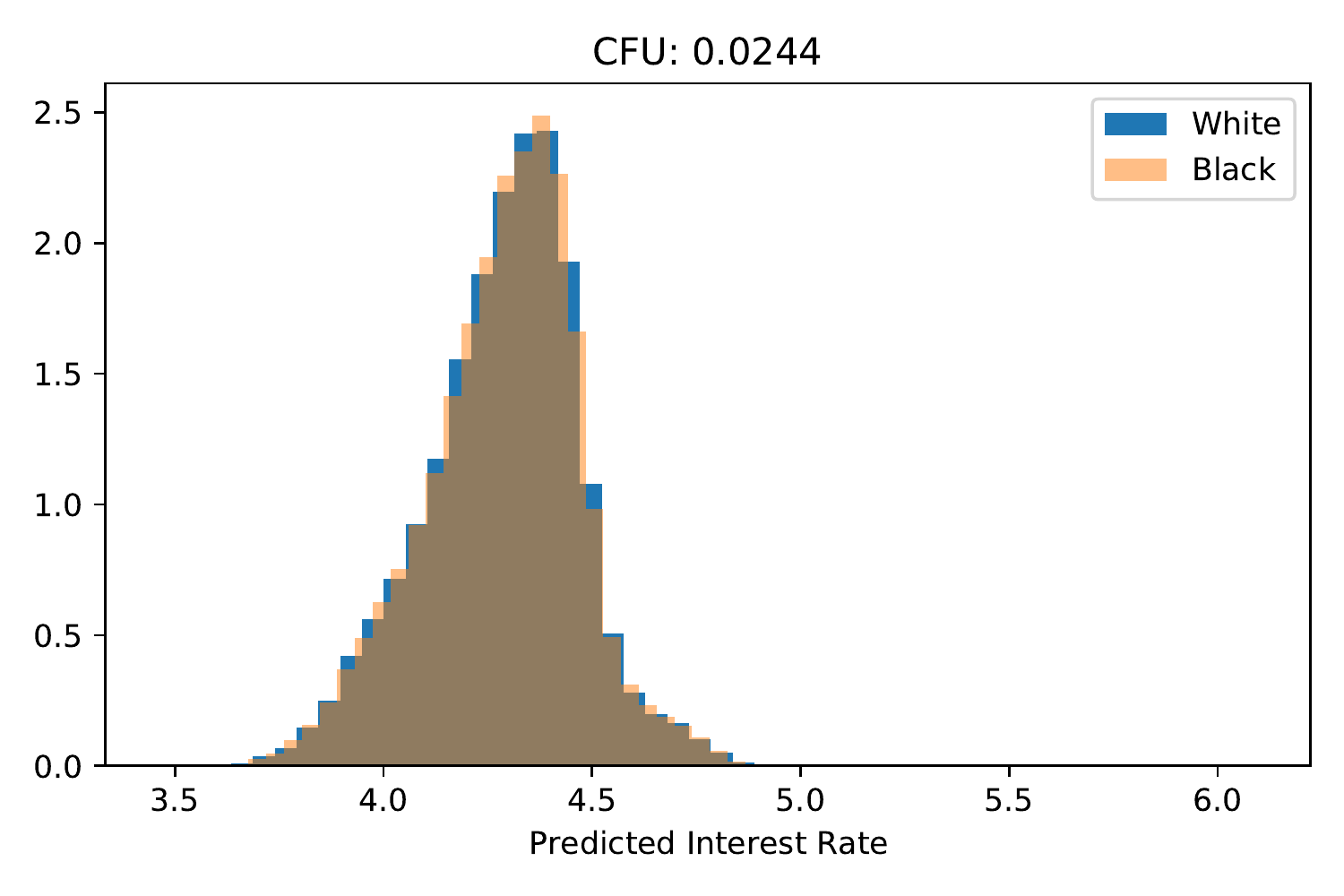}
         \caption{Algorithm 2}
         \label{fig:algo2}
     \end{subfigure}
     \\
     \begin{subfigure}[b]{0.45\textwidth}
         \centering
         \includegraphics[width=\textwidth]{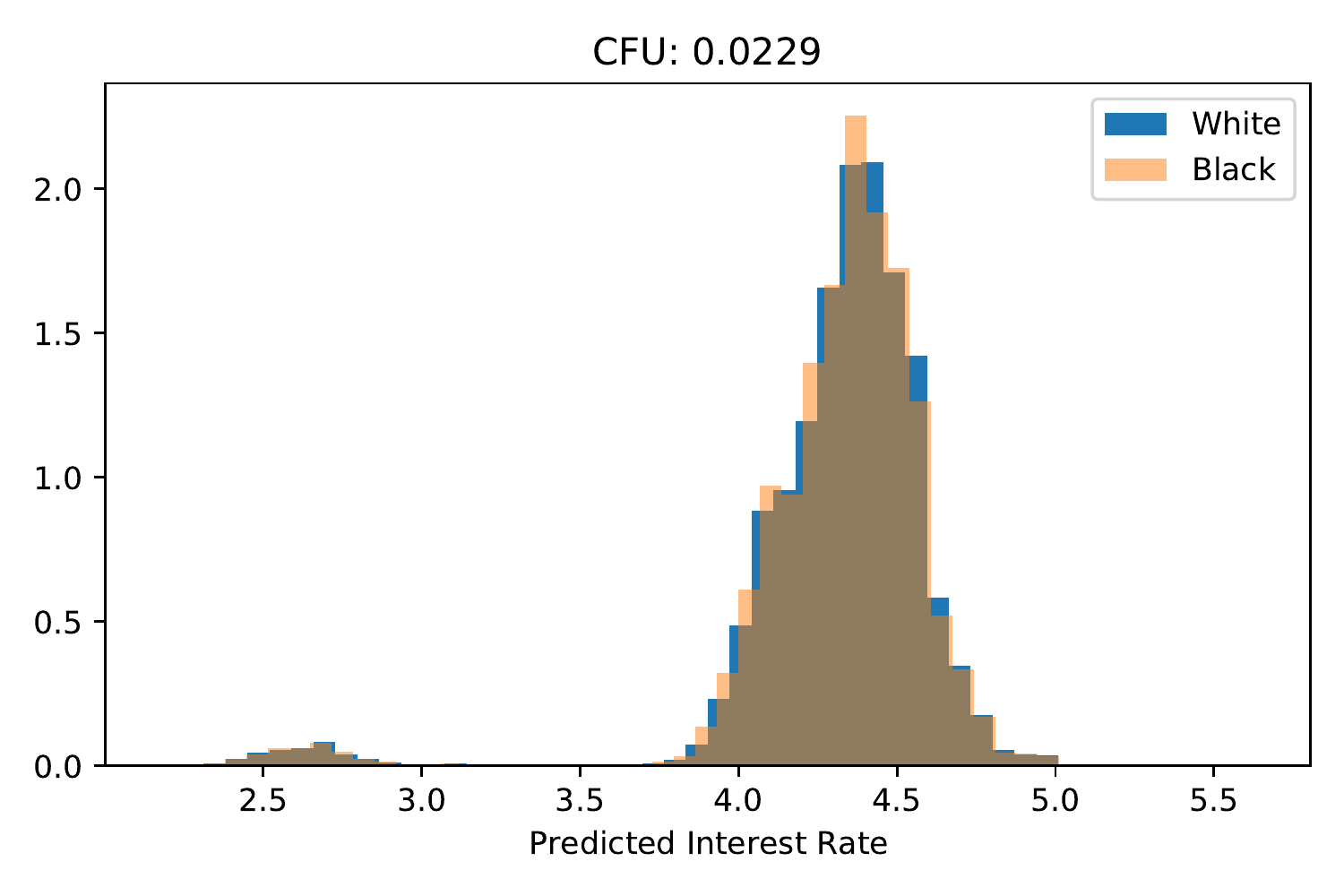}
         \caption{Algorithm 3}
         \label{fig:algo3}
     \end{subfigure}
     \hfill
     \begin{subfigure}[b]{0.45\textwidth}
         \centering
         \includegraphics[width=\textwidth]{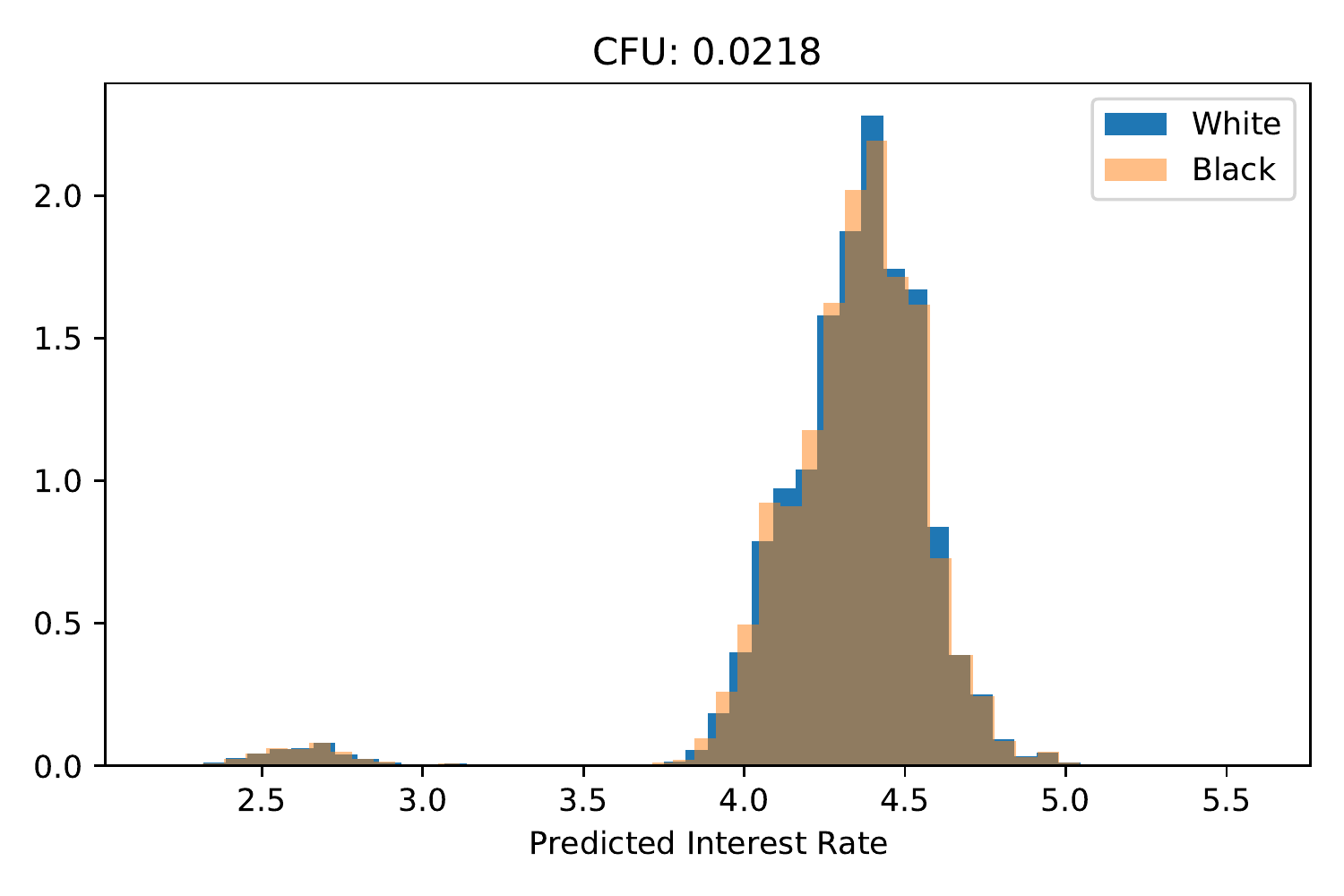}
         \caption{Algorithm 4}
         \label{fig:algo4}
     \end{subfigure}
        \caption{Illustration of the counterfactual fairness in interest rate prediction using linear regression. We show the histograms of the predicted interest rate in the test matched pairs.}
        \label{fig:LR_interest}
\end{figure}

Tables \ref{tab:classification} and \ref{tab:regression} show our detailed evaluation results for the approval and interest rate prediction tasks, respectively. In both tables, Algorithms 2 and 3 show much fairer performance indicating that the main improvement in accuracy comes from removal of the race variable from the feature set. Overall, while we see a mild improvement in CFU using matching in interest rate prediction task (Table \ref{tab:regression}), balancing the dataset does not improve the results over Algorithm 2 in the mortgage approval prediction task (Table \ref{tab:classification}). We also show the detailed histograms for linear regression on the interest rate prediction in Figure \ref{fig:LR_interest}.

Examining the gains in the fairness scores, we see that the biggest decreases in the CFU scores happen because of exclusion of the race variable in both cases. The label randomization in the interest rate prediction further improves the CFU score. Given that we have defined CFUs to have real-world meaning, one may conclude that the unfairness in Algorithms 2, 3, and 4 are negligible. For example, 0.02\% difference in interest rate prediction in real-world is fair, because interest rates are usually in the increments of larger than 0.1\%.

To dive deeper into the results, we also report the accuracy/error metrics for all algorithms in both tasks. Note that randomization does not degrade the accuracy performance of the Algorithm 4. Also, all algorithms are less accurate on the African-American sub-population compared to the non-Hispanic white populations, in both tables. The discrepancy is because our test data is biased and we need to sacrifice the accuracy to avoid the bias existing in the black population and be fair. 
As a final note, we see that none of the algorithms achieve perfect fairness, despite our efforts for avoiding both of group representation and the proxies problems as mentioned in \citep{HOOKER2021100241} as common sources of bias. One possible conclusion is that, eliminating bias in the dataset, even using matching with the fewest assumptions, does not make the machine learning algorithms perfectly fair.

\begin{table}[t]
    \centering
    \caption{Mortgage Approval Prediction Results}
    \label{tab:classification}
    \begin{tabular}{l|l|c|c|c|c}
        \toprule
         & Algorithm & Test CFU & Test AUC & Test AUC /Whites & Test AUC /Blacks   \\
        \midrule
         \multirow{3}{*}{Neural Network}& Algorithm 1& 0.0771 &0.762 &0.783 &0.729 \\
         & Algorithm 2& 0.0224 &0.771 & 0.798 & 0.749\\
         & Algorithm 3& 0.0236 &0.763 &0.787 &0.745 \\
         & Algorithm 4& 0.0249 & 0.768 & 0.794 & 0.749\\
        \midrule
        \multirow{3}{*}{Random Forest}& Algorithm 1& 0.0366&0.781 &0.809 &0.762 \\
         & Algorithm 2& 0.0042&0.783 &0.811 &0.762 \\
         & Algorithm 3&0.0074 &0.772 &0.796 &0.756 \\
         & Algorithm 4& 0.0063 & 0.769 & 0.793 & 0.753\\
        \bottomrule
    \end{tabular}
\end{table}

\section{Summary and Conclusions}
Counterfactual fairness provides an elegant framework for measuring the algorithms. In this paper we propose a randomization procedure added to causal matching to train counterfactually fair algorithms. We empirically show the success of the algorithm in the interest rate prediction in a large mortgage lending data. We also document that simply balancing the data to counterfactual pairs using causal matching does not create perfectly fair machine learning algorithms. 
%The significance of our results is the fact that we provide a concrete evidence for \citep{HOOKER2021100241} with the simplest data and minimum assumptions in balancing. 

% \input{discuss.tex}

\bibliographystyle{apalike}
\bibliography{refs}

\newpage
\clearpage
\appendix
\section{Appendix}
\subsection{Description of the variables}
We provide the description of variables from the HMDA data dictionary\footnote{\url{https://ffiec.cfpb.gov/documentation/2018/lar-data-fields/}}.

\textbf{'interest rate'}: The interest rate for the covered loan or application.

\textbf{'applicant sex'}: Single aggregated sex categorization derived from applicant/borrower and sex fields. The possible values are: Male, Female, Joint, and Sex Not Available.

\textbf{'income'}: The gross annual income, in thousands of dollars, relied on in making the credit decision, or if a credit decision was not made, the gross annual income relied on in processing the application.

\textbf{'applicant race'}: Race of the applicant or borrower. We only use the data corresponding to  Black or African American and Non-Hispanic white. To find the non-Hispanic whites, we combine the race and ethnicity columns in the data.

\textbf{'state code'}: Two-letter state code.

\textbf{'loan type'}: The type of covered loan or application. Possible values are: Conventional (not insured or guaranteed by FHA, VA, RHS, or FSA),  Federal Housing Administration insured (FHA),  Veterans Affairs guaranteed (VA), and USDA Rural Housing Service or Farm Service Agency guaranteed (RHS or FSA).

\textbf{'debt to income ratio'}: The ratio, as a percentage, of the applicant’s or borrower’s total monthly debt to the total monthly income relied on in making the credit decision.

\textbf{'loan to value ratio'}: The ratio of the total amount of debt secured by the property to the value of the property relied on in making the credit decision.

\textbf{'lien status'}: Lien status of the property securing the covered loan, or in the case of an application, proposed to secure the covered loan. The possible values are either Secured by a first lien
or Secured by a subordinate lien.

\subsection{Matching brackets}
\textbf{'income'}:  We use the following brackets: ~ [32,53,107,374].

\textbf{'debt to income ratio'}:  [0, 20, 30, 36, 40, 45, 50, 60].

\textbf{'loan to value ratio'}:  [40, 60, 79, 81, 90, 100].

\end{document}